\documentclass[11pt,letterpaper]{article}
\usepackage[T1]{fontenc}
\usepackage[utf8]{inputenc}
\usepackage{lmodern}
\usepackage[margin=0.8in]{geometry}
\usepackage{microtype}
\usepackage{amsmath,amssymb,mathtools}
\usepackage{graphicx,booktabs,array,multirow}
\usepackage{enumitem}
\usepackage[numbers,sort&compress]{natbib}
\usepackage[font=small,labelfont=bf]{caption}
\usepackage{xcolor}
\definecolor{linkblue}{RGB}{30,80,115}
\usepackage[colorlinks=true,allcolors=linkblue,breaklinks=true]{hyperref}
\usepackage{placeins}

\newcolumntype{L}[1]{>{\raggedright\arraybackslash}p{#1}}
\setlist{itemsep=2pt,topsep=4pt,parsep=0pt}
\makeatletter
\renewcommand{\maketitle}{\begin{center}
  {\LARGE\bfseries\@title\par}\vspace{10pt}
  {\normalsize\@author\par}\vspace{5pt}
  {\small\@date\par}
\end{center}\vspace{5pt}}
\makeatother
\title{Recursive Self-Improvement of Visuomotor Policies through Local Recovery Supervision}
\author{\begin{tabular}{ccc}
Yuzhi Zhang & Xinyu Liu & Yu Zhang\\[3pt]
{\small\texttt{zhangyuzhi@buaa.edu.cn}} &
{\small\texttt{liuxy0109@buaa.edu.cn}} &
{\small\texttt{stevezhang@buaa.edu.cn}}
\end{tabular}\\[5pt]
Beihang University}
\date{October 2026}
\hypersetup{pdftitle={Recursive Self-Improvement of Visuomotor Policies through Local Recovery Supervision},pdfauthor={Yuzhi Zhang, Xinyu Liu, Yu Zhang},pdfsubject={Recursive self-improvement of visuomotor policies with preliminary post-training evaluation}}
\begin{document}
\maketitle
\begin{abstract}
Visuomotor policies can execute familiar tasks yet lack the corrective behavior needed after their own mistakes. We present a framework for recursive self-improvement through local recovery supervision. Each round audits the current policy, generates corrective demonstrations at supported failure states, and uses them to update the policy that drives the next round of collection. An offline auditor locates unresolved failures using coarse and dense temporal evidence and specifies observable repair goals. A fixed multimodal agent acts as a tool-using teacher, generating recovery actions through observation, computation, execution, and feedback. The frozen student tests whether each teacher endpoint supports further progress. If continuation fails, the system restores that endpoint and extends the demonstration. Action-level quality assessment then defines continuous training windows with aligned observations, quality weights, and validity masks. Only the student is deployed. In a preliminary LIBERO-Goal study, recovery-augmented post-training achieves 88 successful episodes out of 100 validation scenes, compared with 78 for original-data continuation from the same $\pi_0$ checkpoint. An earlier BC-RNN study on robomimic Can improves success from 102/130 to 112/130 using 26 local recovery segments. Both comparisons match 2,000 additional optimization steps.
\end{abstract}

\section{Introduction}
\label{sec:intro}
Visuomotor policies map visual observations and robot state to actions, optionally conditioned on a task instruction. Large demonstration datasets and generalist policies have expanded the range of tasks that can be learned with a shared model~\cite{bridge,oxe,octo,openvla,pi0,pi05}. Action chunking and generative policies further improve the representation of manipulation behavior~\cite{act,diffusionpolicy}. Yet knowing how to complete a task does not imply knowing how to recover when execution deviates from the demonstrations. Recent recovery benchmarks make this gap explicit by evaluating policies from intermediate failure states~\cite{roborecover,liberorecover}. After missing a grasp, a policy may continue along a familiar transport motion with an empty gripper. The remaining task is recognizable, but the behavior needed to make its existing competence useful again is missing.

Failures expose this gap in a form that can guide further training. Interactive imitation learning queries expertise on states reached by the learner~\cite{dagger,dart}; human interventions provide corrections when autonomous execution needs help~\cite{hgdagger,iwr,thriftydagger,sirius}. Recent work extends this principle through recovery learning, automated intervention generation, and policy post-training~\cite{rac,intervengen,recap,sop}. Our question is how an automatic teacher should turn the current visuomotor policy's failures into useful action supervision. Completing the task is one possible response, but it does not determine when a demonstration should start, where it should end, or which of its actions deserve imitation.

This question becomes central when collection and learning repeat. Updating a policy changes the states it visits, the mistakes it makes, and the corrections it needs. A fixed recovery dataset may therefore cease to target its weaknesses. We formulate recursive self-improvement (RSI) as a policy-driven cycle: audit the current student, collect local recovery supervision, update its parameters, and audit the resulting policy again. The recursion lies in this feedback between policy behavior and subsequent training data. The multimodal teacher supplies additional competence while remaining fixed; the student determines both where correction is needed and whether a demonstration has gone far enough.

Consider a failed grasp. The final empty-gripper image reveals an unsuccessful outcome, but an earlier state may offer a better opportunity to correct it. Once the teacher secures the object, the immediate failure is resolved; the student may nevertheless release it before reaching the destination. The teacher then needs to demonstrate more than grasp acquisition. Conversely, requiring the teacher to finish every remaining step can spend collection effort on behavior the student already performs reliably. Even an accepted recovery may contain avoidable detours. A useful collection procedure must make these decisions separately.

\begin{figure}[t]
\centering
\includegraphics[width=\linewidth]{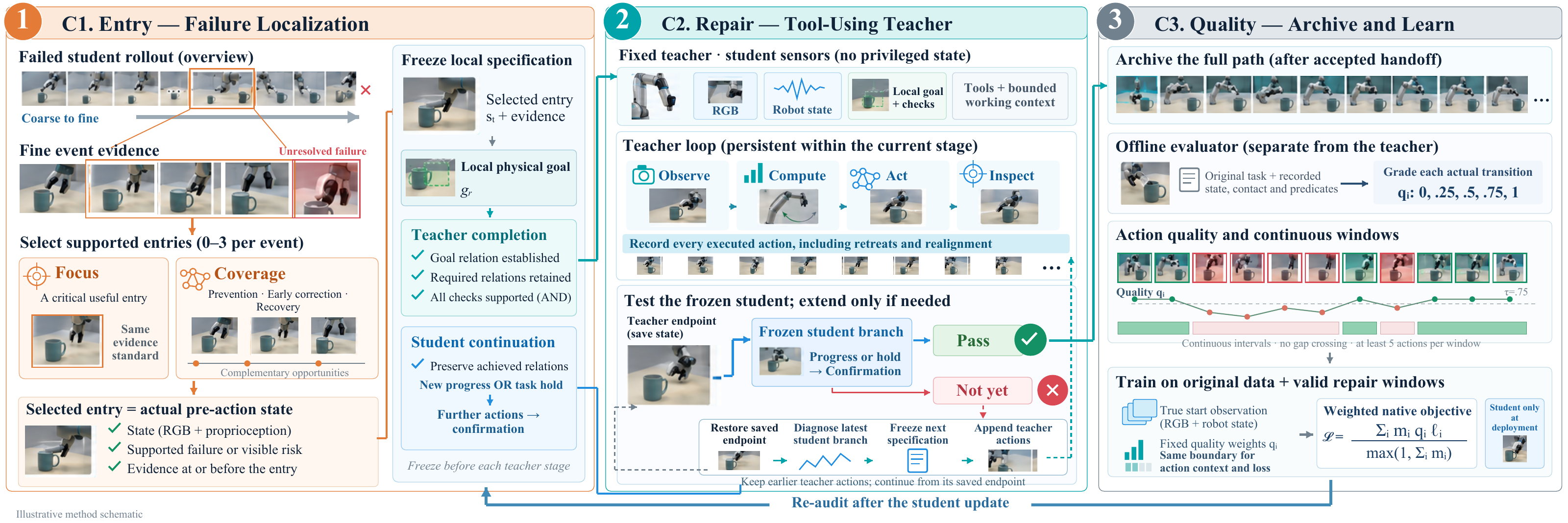}
\caption{\textbf{Recursive policy improvement through local recovery supervision.} Evidence determines where to intervene; execution by the frozen student determines how far the teacher must demonstrate; transition quality determines which continuous intervals are used for training. Failed student branches provide diagnostic evidence, while only accepted teacher actions supply recovery labels. After post-training, the updated student supplies the next audit, closing the outer loop. Teacher weights remain fixed. Robot thumbnails are synthetic method illustrations.}
\label{fig:overview}
\end{figure}

We organize local recovery supervision around three boundaries: the entry supported by failure evidence, the endpoint supported by student execution, and the continuous intervals supported by action quality (Fig.~\ref{fig:overview}). First, an offline auditor examines a coarse rollout and then dense evidence around an unresolved event. Focus prioritizes a critical intervention, whereas Coverage seeks complementary prevention, early-correction, and recovery opportunities. The selected state is paired with a prospective specification: the physical relation to establish, the observations needed to confirm it, and achieved progress to preserve.

A tool-using teacher generates a correction through observation, computation, bounded action, and inspection of the resulting scene. It can compute a new target or write a short program, but it must justify completion from observable physical evidence. The frozen student then executes from the teacher endpoint. If it cannot make and sustain useful progress, the system discards that diagnostic branch, restores the teacher endpoint, and appends a further correction. This makes demonstration length depend on the recipient's behavior rather than on teacher confidence alone.

After a successful handoff, an offline evaluator grades each executed teacher action in the state from which it was actually taken. Maximal continuous intervals above a quality threshold become training windows. Their observations, targets, action context, and loss masks remain aligned; actions on student test branches never become imitation targets. Mixing these windows with the original data updates the student without retaining a teacher at deployment.

The contribution is a concrete procedure for recursive improvement of visuomotor policies: evidence-based entry selection and recipient-driven endpoint extension; a closed-loop tool teacher governed by local physical specifications; and continuous, quality-weighted recovery supervision for post-training. We describe the implementation and report a $\pi_0$ study on a custom LIBERO-Goal split, together with an earlier BC-RNN experiment on robomimic Can. With 2,000 additional student updates in each comparison, recovery supervision raises success from 78/100 to 88/100 for $\pi_0$ and from 102/130 to 112/130 for BC-RNN.

\section{Related Work}
\label{sec:related}
\paragraph{Corrective imitation and post-training.}
DAgger makes the learner's state distribution central to imitation learning~\cite{dagger}. DART exposes demonstrations to controlled perturbations, and HG-DAgger, intervention weighted regression, ThriftyDAgger, and SIRIUS organize when experts intervene and how their behavior enters training~\cite{dart,hgdagger,iwr,thriftydagger,sirius}. RaC scales recovery and correction data for long-horizon tasks~\cite{rac}. RECAP combines experience and interventions through advantage-conditioned policy learning, while SOP provides an online system for collecting experience and updating generalist policies~\cite{recap,sop}. We construct local corrections for states poorly covered by the student's demonstrations.

\paragraph{Self-improvement in robot learning.}
RoboCat uses trained policies to generate experience for subsequent training iterations~\cite{robocat}. EmbodiRSI combines agent-assisted corrections and adaptive data collection in task-specific simulations to improve policies before transfer to physical robots~\cite{embodirsi}. These works establish the broader feedback loop between policy capability and data acquisition. Our focus is how to construct local recovery supervision within that loop: failure evidence supports the entry, execution by the recipient policy supports the endpoint, and action quality determines the continuous intervals used for learning. The recursive setting motivates fresh audits, while these supervision decisions define the method studied here.

\paragraph{Learning recovery from failures.}
RACER augments demonstrations with recovery trajectories and rich language annotations, then uses a vision-language supervisor to guide a visuomotor actor~\cite{racer}. FLARE teaches retries through perturbed demonstrations and learns reset skills for failures that disrupt the environment; an online monitor chooses between task execution and reset~\cite{flare}. These methods connect recovery supervision with structured guidance during execution. Our teacher instead operates during data collection, and the resulting corrections are absorbed into a student that deploys alone. Fail2Progress addresses a related data problem by generating simulation environments resembling observed failures and adapting a skill effect model~\cite{fail2progress}. We target the actions needed to recover from a particular student's execution state.

\paragraph{Recovery-data generation and continuation.}
MimicGen and SkillMimicGen expand demonstrations through task structure and motion transfer~\cite{mimicgen,skillgen}; IntervenGen generates corrective interventions from a small human seed set~\cite{intervengen}. WM-DAgger and Dream2Fix use predictive or counterfactual models to acquire corrective experience~\cite{wmdagger,dream2fix}. Kintsugi-VLA repeatedly restores failed-rollout states and probes a privileged expert to estimate recoverability, which guides recovery-start selection~\cite{kintsugi}. MAGMA-GEN validates proposed corrections by re-execution and downstream continuation, producing supervision over task-level tool calls~\cite{magma_gen}. Here, the frozen recipient policy tests the endpoint of an executed native-action demonstration. A failed check causes the teacher to extend its saved path, while the student branch supplies diagnostic evidence rather than training labels.

\paragraph{Failure evidence and robot agents.}
ProTracer uses proprioceptive changes and visual evidence to diagnose failures and localize their onset~\cite{protracer}. Sentinel combines monitoring of action consistency with reasoning about task progress, and ViFailBack uses visual symbols for diagnosis and correction~\cite{sentinel,vifailback}. We use temporal evidence to justify an actionable intervention state, which need not coincide with the first anomaly. For execution, SayCan, Code as Policies, and VoxPoser connect language reasoning to skills, programs, and spatial representations~\cite{saycan,codeaspolicies,voxposer}. AERIS and AeroWeaver also connect language-agent decisions to closed-loop robotic execution~\cite{lou2026aeris,lou2026aeroweaver}. Agent as Policy keeps a tool-using agent in the execution loop and provides the client underlying our control bridge~\cite{agp}. We adapt this execution interface to local recovery collection with fixed specifications, completion evidence, saved endpoints, and student handoff checks.

\paragraph{Quality of imitation targets.}
SSDF scores imperfect trajectory segments using learned similarity, filters low-quality data, and applies weighted behavioral cloning~\cite{ssdf}. We extend this perspective to supervision extracted from an extended recovery path. An action is assessed in the state actually reached; a necessary retreat after a poor approach can be useful even when the approach should be excluded. Retained transitions form continuous intervals with their true initial observations. For joint action-chunk prediction, validity also controls future action positions supplied as model input, in addition to the supervised loss. This preserves a distinction between accepting a handoff, archiving an executed path, and choosing its imitation targets.

\paragraph{Policy representations and adaptation.}
OpenVLA-OFT studies how action decoding, chunking, and the training objective affect VLA adaptation~\cite{oft}. SpatialVLA incorporates explicit spatial representations, while GeoPredict adds predictive kinematic and geometric supervision~\cite{spatialvla,geopredict}. Cosmos Policy adapts a pretrained video model to generate actions, future observations, and values~\cite{cosmospolicy}. These approaches improve how policies represent and learn manipulation behavior. Local recovery collection addresses the complementary question of which corrective experience to provide. Applying that experience across architectures requires preserving each model's action representation and objective, including its treatment of invalid action positions. This motivates comparisons within each student architecture under matched training conditions.

\paragraph{Evaluating recovery.}
RoboRecover evaluates policies from execution-induced deviation states reconstructed by replay~\cite{roborecover}. LIBERO-RECOVER organizes failures by the correction they require, from action retries to changes in object and environment state~\cite{liberorecover}. Both make recovery an explicit evaluation target beyond completion from nominal starts. This distinction is central to our setting: better task success after post-training and better recovery from a given failure are related but separate outcomes. The evaluation protocol therefore includes both nominal episodes and held-out failure states, with the original task unchanged.

\section{Recursive Improvement through Local Recovery Supervision}
\label{sec:method}

\subsection{Problem setup and data flow}
Let $g$ denote a task, $s_t$ the environment state, and $x_t$ the student's visual and proprioceptive observation. A visuomotor policy $\pi_\theta(a_t\mid x_{\leq t},g)$ maps its available observation history to robot actions; the dependence on $g$ is implicit for a single-task policy and may be an explicit language instruction for a generalist policy. The observation need not reveal the full environment state. Starting from a student trained on demonstrations $\mathcal D_0$, each collection round freezes its current parameters and executes it in training scenes. A failed rollout is
\begin{equation}
\tau=(s_0,a_0,s_1,\ldots,a_{N-1},s_N).
\end{equation}
An intervention at $0\leq t<N$ restores the state immediately before $a_t$. A tool teacher then generates a new physical branch,
\begin{equation}
\tau_T=(y_0=s_t,u_0,y_1,\ldots,u_{K-1},y_K),
\label{eq:teacher_path}
\end{equation}
where $u_i$ is an executed robot action. Subsequent teacher states generally differ from those in the failed rollout.

The method assigns three different roles to trajectory boundaries. Failure evidence determines where a repair starts. Actual student execution determines whether the teacher has reached a useful handoff. Action quality determines which continuous intervals become training examples. The student remains fixed during collection and is updated only after the accepted repairs have been scored and converted into training windows.

\subsection{Locate a failure and justify an intervention}
\label{sec:event}
\paragraph{From an unresolved event to a state.}
The localization model first reads a coarse view of the failed rollout. Uniform samples, approximately one per second, establish task context. Additional samples around motion starts, stops, direction changes, and gripper changes expose transitions that merit closer inspection. These changes propose evidence locations; they do not by themselves identify a failure. Offline diagnosis can also use recorded physical-state summaries that are withheld from the acting teacher and deployed student.

The model describes one unresolved event through three elements: the expected task relation, the observed discrepancy, and evidence that subsequent student behavior does not resolve it. For example, a closed gripper moving away while the object remains on the table supports an empty grasp. A brief pause is insufficient evidence for a stall; a stall requires sustained ineffective behavior. The description includes an interval for further inspection.

Dense inspection then distinguishes the last supported normal state, uncertain onset, first sufficient evidence, and later confirmation. Let $E_{\mathcal U}$ and $E_{\mathcal W}$ denote indexed evidence in the overview and local window, respectively. The two model calls produce
\begin{equation}
\begin{aligned}
(\hat e,I_e)&=G_{\mathrm{coarse}}(g,E_{\mathcal U}),\\
\mathcal T_m&=G_{\mathrm{fine}}(g,\hat e,E_{\mathcal U},E_{\mathcal W};m),
\end{aligned}
\label{eq:localization}
\end{equation}
where $m$ selects Focus or Coverage. Each returned entry includes a time $t$, the current problem, an intervention rationale, evidence citations, and the distinct supervision it would provide. The implementation checks that entries are supplied local frames, that no two entries coincide, and that each rationale cites evidence no later than its entry. A nonempty set has exactly one primary entry and at most three entries in total. Localization is offline: the model can inspect later context, although an entry must have a justification grounded at or before $t$.

\paragraph{Focus and Coverage.}
Focus prioritizes the earliest supported state at which a useful correction remains possible. It usually yields one entry; additional entries require a different corrective need. Coverage seeks complementary opportunities for prevention, early correction, and recovery within the same event. Prevention requires an observable risk at that time. Both modes share the evidence requirements, output checks, and downstream procedure, and either may return zero to three entries. They differ only in the selection preference, not in the meaning of a valid intervention.

In the current implementation, the overview includes at most eight change centers. The local interval extends approximately $0.5$ seconds beyond the proposed event on each side and contains up to about $96$ samples. One additional targeted inspection is allowed. If the evidence remains insufficient, the model abstains. Each accepted entry starts a separate repair attempt from its actual pre-action state.

\subsection{Freeze a local repair specification}
\label{sec:spec}
\begin{figure}[t]
\centering
\includegraphics[width=\linewidth]{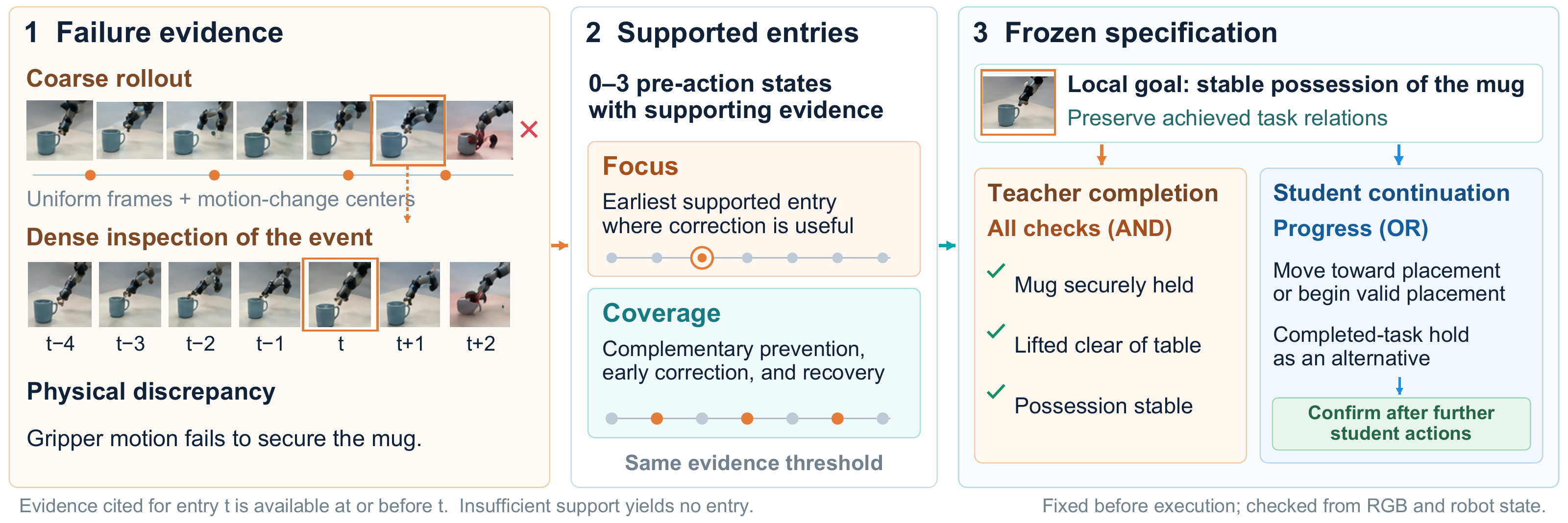}
\caption{\textbf{Failure evidence and prospective repair conditions.} The auditor separates an unresolved physical discrepancy from ordinary adjustment, then selects supported pre-action states. Focus and Coverage use the same evidence standard. A frozen specification assigns different criteria to teacher completion and student continuation. Robot thumbnails are synthetic illustrations of the mechanism, not recorded experimental frames.}
\label{fig:localization}
\end{figure}
Before the teacher acts, a specification model defines the nearest useful physical outcome that removes the current obstacle while protecting prior task progress. A target such as stable possession requires evidence of a held object, rather than only an approach or a closed-gripper command. The goal limits the required repair without prohibiting further correct progress.

For teacher stage $r$, the specification is
\begin{equation}
\Gamma_r=(g_r,\mathcal E_r,b_r;\mathcal P_r,f_r,c_r),
\label{eq:spec}
\end{equation}
where $g_r$ is the local goal, $\mathcal E_r$ contains its completion checks, and $b_r$ describes task relations to preserve. The student fields specify alternative early progress conditions $\mathcal P_r$, a completed-task alternative $f_r$, and confirmation conditions $c_r$.

Teacher completion uses one to three checks with an AND relation. Each states a physical claim, its observable supporting evidence, and the nearest plausible false positive. A possession check, for instance, distinguishes stable object motion with the gripper from a closed gripper moving alone. Checks must use sensors actually available to the teacher.

Student progress uses an OR relation: one supported alternative can trigger confirmation. A nonterminal specification includes one or two early progress conditions and one completed-task alternative. Preservation concerns achieved task relations, not incidental poses. Replacing a grasp by a correct placement can preserve task progress even though possession ends. If the task is already complete, stable completion can be the sole alternative.

The specification is structured natural language with explicit identifiers and evidence fields. It is frozen before each stage. Independent attempts from the same entry reuse the initial specification. After a failed handoff, a new specification governs only the additional teacher stage; it does not revise the verdict on the earlier student branch. The teacher receives the local goal and completion fields, while the continuation observer receives the student criteria.

\subsection{Generate executed recovery actions with a tool teacher}
\label{sec:teacher}
\paragraph{Observation, computation, action, and feedback.}
A fixed multimodal agent serves as the teacher and controls the robot through a persistent tool session. It reads the original task, the active teacher specification $\Gamma_r^T=(g_r,\mathcal E_r,b_r)$, interface documentation, and remaining budget. Within the session, it inspects observations, computes geometric targets or writes a short program, executes bounded commands, and examines the returned feedback. For interaction round $\ell$,
\begin{equation}
\begin{aligned}
p_\ell &= T_\phi(g,\Gamma_r^T,h_\ell),\\
(U_\ell,F_\ell)&=\operatorname{Execute}(p_\ell),\\
h_{\ell+1}&=h_\ell\oplus(p_\ell,U_\ell,F_\ell),
\end{aligned}
\label{eq:tool_teacher}
\end{equation}
where $\phi$ is fixed, $h_\ell$ is the session history, $p_\ell$ is a command or program, $U_\ell$ contains the executed native actions, and $F_\ell$ contains observations and tool feedback. One model turn may issue several serial requests. Only executed robot actions become imitation targets; code and observation requests do not.

The interface exposes state and image queries, native actions, end-effector motion, gripper commands, completion, and abort. Motion helpers specify coordinates, units, limits, and geometric control. They do not select a task-specific recovery script. Following an unsuccessful grasp, the teacher can reopen, retreat, recompute an alignment, and inspect whether the object moves with the gripper. The goal fixes the required physical result, while feedback determines the next action. The request client is derived from Agent as Policy~\cite{agp}; specification handling, the simulation bridge, handoff tests, and trajectory construction implement the procedure described here.

\paragraph{Completion and information access.}
The teacher receives the same prescribed RGB views and robot-state channels as the student, with the same deterministic image preprocessing. Hidden object states, depth, contacts, and task oracles are not exposed through its control interface. It additionally receives the local specification, interface documentation, and its own interaction history. Sensor alignment therefore does not imply identical total information.

To finish a stage, the teacher cites the active specification, every completion check, and current observation records. The host verifies identifiers, evidence-channel coverage, and output structure. The teacher makes the semantic judgment that the required physical relations hold; there is no second local-success oracle. A commanded pose or servo convergence is insufficient evidence by itself. All executed actions are retained, including retreats and unsuccessful adjustments.

Each stage starts a new session. An extension may receive the previous stage's last six commands and associated student-visible observations. Independent attempts start fresh, with neither updated teacher weights nor a cross-attempt memory. A failed stage or a completion claim with no new actions terminates the attempt.

The current simulation bridge bounds a native-action request to 1--40 steps. Physics advances only while executing action requests; model inference, image inspection, and program generation consume wall-clock time without advancing the simulated scene. This scheduling isolates bounded physical execution from model latency. A continuously evolving physical platform would require an explicit treatment of that latency.

\subsection{Test the student and extend the teacher path}
\label{sec:handoff}
\begin{figure}[t]
\centering
\includegraphics[width=\linewidth]{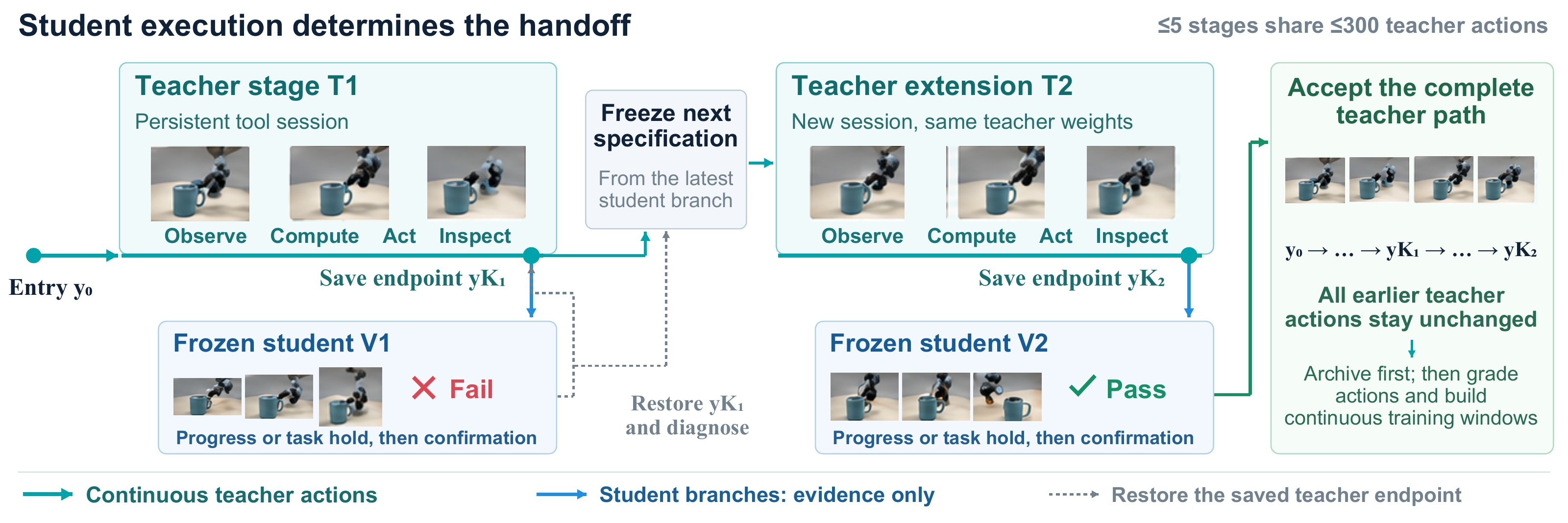}
\caption{\textbf{Branch, restore, and extend.} Student tests branch from saved teacher endpoints. A failed test supplies diagnostic evidence for a new specification; the next teacher stage resumes from the saved endpoint, preserving the earlier teacher path. Student test and replay actions provide no imitation targets. Robot thumbnails are synthetic method illustrations.}
\label{fig:handoff}
\end{figure}
Local repair completion is followed by an actual test of the frozen student. We save the teacher endpoint $y_{K_r}$ and initialize a student branch $V_r$ there. With stage lengths $k_r>0$ and $K_r=\sum_{j=1}^r k_j$,
\begin{equation}
\begin{aligned}
y_0&\xrightarrow{T_1}y_{K_1}\xrightarrow{T_2}\cdots\xrightarrow{T_R}y_{K_R},\\
V_r&=(z_0^{(r)}=y_{K_r},v_0^{(r)},\ldots,z_{M_r}^{(r)}).
\end{aligned}
\label{eq:branches}
\end{equation}
The first line is the continuous teacher path. Student branches in the second line supply acceptance and diagnostic evidence, never imitation labels. Pending student action chunks are cleared at handoff; any required causal history follows the actual audit prefix and teacher path.

\paragraph{Progress followed by confirmation.}
The observer inspects the student every $\Delta$ actions for at most $H_S$ actions. It checks whether the student preserves the teacher's achieved task relations and makes meaningful progress satisfying an alternative in $\mathcal P_r$. Mere motion or a reduced position error is insufficient unless the frozen criterion identifies it as useful evidence. The first supported advance establishes a progress anchor. The student then executes $C$ more actions, and that advance must remain valid or develop into further task progress.

Let $B_r(V)$ mean that the required prior relations remain valid. Define $A_r(V)$ as the OR of the early-progress alternatives in $\mathcal P_r$ and the completed-task alternative $f_r$, and let $R_r(V^+)$ mean that the resulting progress or completion survives confirmation. The acceptance rule is
\begin{equation}
\operatorname{Accept}_r=\operatorname{ReceiptValid}_r
\land B_r(V_r)\land A_r(V_r)
\land B_r(V_r^+)\land R_r(V_r^+).
\label{eq:acceptance}
\end{equation}
Here $V_r^+$ includes the confirmation continuation. The observer distinguishes the teacher's achievement at release from the student's newly established advance. Before an advance is established, withdrawing a tentative movement is not automatically a loss of protected progress. Once confirmed progress is sought, the new anchor must be preserved or legitimately advanced. Uncertain evidence does not pass, and destruction of a protected relation can end the test early.

If the original task is complete, the alternative $f_r$ requires completion to hold at at least two post-action observations, including the current endpoint, followed by confirmation. Thus, a completed task need not exhibit unnecessary extra movement. Our implementation uses $(H_S,\Delta,C)=(60,10,10)$; progress first detected at action $60$ is checked through action $70$. Acceptance supplies evidence of short-term handoff usability, rather than a guarantee of eventual full-task success.

\paragraph{Restore the teacher endpoint before extending.}
When continuation fails, we save the student branch for diagnosis and restore its starting teacher endpoint. The specification model uses that endpoint and the latest student behavior to define the next useful condition. The teacher appends actions from the restored endpoint, preserving all earlier teacher actions and the original entry $t$. If the student dropped an object after handoff, the extension therefore starts from the saved state where the teacher still held it, rather than from the student's damaged state.

The simulator reconstructs endpoints by replaying the audit prefix and accumulated teacher actions, and verifies physical and controller state. An unsuccessful restoration ends the attempt. The current limits are five stages, $300$ teacher actions, and $7{,}200$ seconds per attempt, shared across stages. A teacher session permits at most $100$ robot requests and $1{,}200$ seconds within that remaining budget. Local completion at the action cap still triggers a student test, but a failed test cannot start another extension. A pass accepts the entire accumulated teacher path once; failed attempts contribute no training labels.

\subsection{Score actions and retain continuous training intervals}
\label{sec:quality}
An accepted handoff does not make every preceding action a good target. An offline evaluator therefore scores each executed transition after the complete teacher path has been accepted. The question is whether $u_i$, from its actual preceding state $y_i$, advances the original task, preserves a needed relation, or establishes a useful prerequisite, and whether it has an avoidable defect. A necessary return after an earlier detour can be useful even when the detour itself is not.

The evaluator reads the original task, action semantics, recorded transitions, object and joint relations, contacts, and official task predicates. In the LIBERO implementation, these semantics are obtained from benchmark definitions. The teacher's local goal is excluded from scoring, so satisfying an irrelevant intermediate target cannot define task utility. Unlike the acting teacher, this offline evaluator may use privileged simulation records. The scores describe task utility under recorded evidence, not calibrated success probabilities or guarantees that every action is inferable from the student's restricted input.

Each action receives one fixed ordinal grade,
\begin{equation}
q_i\in\{0,0.25,0.5,0.75,1\},
\end{equation}
corresponding to harmful, ineffective, useful with material defects, useful with minor defects, and effective actions. Insufficient evidence receives weight zero while retaining an explicit unknown status. A failed scoring request remains pending rather than becoming a semantic judgment. Actions are scored once, and overlapping windows reuse their grades.

We preserve the complete accepted archive and extract maximal continuous intervals $[s,e)$ with $q_i\geq\eta$, using $\eta=0.75$. A lower or unknown score breaks the interval. Separated good actions are never concatenated across a physical gap. With prediction horizon $H_A$ and minimum usable prefix $L_{\min}$, a valid anchor and its length satisfy
\begin{equation}
j\in\{s,\ldots,e-L_{\min}\},\qquad
h_j=\min(H_A,e-j).
\label{eq:windows}
\end{equation}
Its training input is the true student observation at $y_j$, paired with $u_j,\ldots,u_{j+h_j-1}$. In particular, later good actions are not assigned to the image at the original intervention. For $H_A=50$ and $L_{\min}=5$, a $25$-action interval yields $21$ anchors. Its last four actions cannot begin a new window, but remain targets in earlier windows. Windows may cross verified teacher-stage boundaries when the physical path and quality interval are continuous.

\subsection{Post-train the student with valid action context}
\label{sec:learning}
Recovery windows use $H_A$ tensor positions with an explicit validity mask $m_{b,k}$. Positions beyond $h_j$ are cleared before and after action transformations. In our $\pi_0$ adapter~\cite{pi0}, the mask controls action targets, noisy action inputs, flow targets, attention queries and keys, and loss. Loss masking alone would still allow excluded noisy action tokens to influence valid positions. The original demonstrations retain the baseline's tail-padding protocol, full validity, and unit weights.

Original and recovery windows are pooled and uniformly shuffled without recovery oversampling. If $\ell_{b,k}(\theta)$ is the student's native per-position error averaged over action dimensions, the objective is
\begin{equation}
\mathcal L(\theta)=
\frac{\sum_{b,k}m_{b,k}q_{b,k}\ell_{b,k}(\theta)}
{\max(1,\sum_{b,k}m_{b,k})}.
\label{eq:loss}
\end{equation}
The denominator counts valid targets rather than summing their quality weights. Lower grades therefore attenuate their loss contributions instead of being renormalized to full strength. Quality weights are fixed metadata, and gradients do not pass through the evaluator. For $\pi_0$, the native flow-matching target and architecture remain unchanged apart from validity handling.

Deployment runs the visuomotor student alone with its normal input and action-generation protocol. It does not read teacher specifications, quality scores, or future valid lengths. The current implementation and empirical studies use simulation with verified state restoration; transfer to another platform requires an observation and control adapter, trustworthy restoration or reset procedures, and appropriate offline task evidence.

\subsection{Close the recursive improvement loop}
\label{sec:rsi}
Let $k$ index policy updates, distinct from the teacher stages $r$ within a repair. The outer procedure is
\begin{equation}
\begin{aligned}
\mathcal F_k &= \operatorname{Audit}(\pi_{\theta_k},\mathcal S_{\mathrm{train}}),\\
\mathcal W_k &= \operatorname{CollectRecovery}(\mathcal F_k,\pi_{\theta_k},T_\phi),\\
\theta_{k+1} &= \operatorname{PostTrain}(\theta_k;\mathcal D_0,\mathcal W_k).
\end{aligned}
\label{eq:rsi}
\end{equation}
Here $\mathcal F_k$ contains failed rollouts on collection scenes $\mathcal S_{\mathrm{train}}$. Collection includes entry localization, teacher execution, student continuation checks, and quality assessment, producing the valid windows $\mathcal W_k$. Post-training applies the student's native objective with the validity and weighting rules above. The teacher parameters $\phi$ remain fixed across rounds.

The updated policy supplies the next audit and all subsequent handoff checks. Thus both the failures selected for repair and the amount of demonstration needed can change with the student's competence. Endpoint extension refines one demonstration without changing student weights; recursive post-training changes the policy that determines later supervision. This loop defines a procedure for improvement, not a guarantee of monotonic gains. The reported experiments evaluate one post-training update; Sec.~\ref{sec:future} specifies the controls needed to evaluate repeated rounds.

\section{Preliminary Empirical Studies}
\label{sec:empirical}

We evaluate the learning step of the proposed RSI framework: whether a small amount of recovery supervision improves a visuomotor student beyond further training on its original demonstrations. The main study uses $\pi_0$ on LIBERO-Goal. An earlier BC-RNN experiment on robomimic Can tests the same data principle with a smaller recurrent policy. Within each study, both post-training conditions start from identical student parameters and receive the same number of additional optimization updates. Each comparison covers one student update, rather than a complete multi-round evaluation.

\subsection{$\pi_0$ on LIBERO-Goal}
\label{sec:empirical_protocol}

\paragraph{Data and initial policy.}
We use all ten LIBERO-Goal tasks~\cite{libero}, built on robosuite~\cite{robosuite}, with 50 original demonstrations per task. A fixed split with seed 7 assigns 30 scenes per task to training, ten to validation, and ten to a reserved test set. Scenes are grouped by task, scene XML, and initial physical state, with the simulation clock removed and numerical values rounded to $10^{-6}$. The three splits have disjoint scene identifiers. Normalization statistics are computed from training data only.

The simulation adapter uses seven-dimensional pose-and-gripper actions at 20\,Hz. Inputs contain an external and a wrist RGB view plus eight robot-state values: end-effector position, axis-angle orientation, and two gripper joints. Both teacher and student receive the same deterministic padding and bilinear resizing of the $128\!\times\!128$ images to $224\!\times\!224$. Original dataset records were checked against replay for temporal alignment: their post-action observation at index $i$ is paired with action $i+1$, and the final unmatched frame is dropped. Recovery records directly store pre-action observations. Original windows retain the baseline's repeated-tail padding; validity masking applies to recovery windows.

The student starts from the public $\pi_0$ base model~\cite{pi0}. Initial training runs for 5,000 updates, with validation every 1,000 updates. The corresponding success rates are 24\%, 73\%, 77\%, 75\%, and 76\%; we select the 3,000-update checkpoint as $M_0$. This selection uses the same validation scenes reported below. The reserved test split has not been evaluated in this study.

\paragraph{Recovery supervision.}
We audit the frozen $M_0$ on 300 training scenes, obtaining 233 successes and 67 failures. Recovery collection uses Focus selection and a tool teacher configured as \texttt{gpt-6.1-sol} with medium reasoning effort. It yields 27 accepted teacher paths containing 3,974 native actions. The collection was assembled during system development: retry limits changed from three attempts per entry to one, and accepted paths span successive prompt and code revisions. We therefore report the size of the resulting dataset, rather than treating its acquisition yield as a fixed-protocol teacher success rate. All paths are graded with the same final action-quality procedure before training.

With a quality threshold of 0.75 and a minimum valid run length of five, the resulting training view covers 2,095 distinct teacher actions in 114 contiguous runs. These produce 1,639 overlapping recovery windows of length 5--50, with a mean of 17.48 valid actions. The windows contain 28,649 valid target occurrences; overlap means that these are not distinct physical actions. Combined with 37,907 original windows, recovery windows account for 4.14\% of training anchors and 1.49\% of valid target occurrences. We shuffle this mixture without recovery oversampling. Student continuation branches contribute no action targets.

\paragraph{Matched post-training.}
Both conditions load the same $M_0$ parameters and run 2,000 additional updates. Ordinary continuation uses only the original demonstrations; recovery continuation uses the original data together with quality-bounded recovery windows. Both restart AdamW and the learning-rate schedule, use seed 7 and a batch size of 48 on three RTX A6000 GPUs, and are evaluated at their final update. The learning rate warms up for 200 updates to $2.5\!\times\!10^{-5}$ and decays by a cosine schedule to $2.5\!\times\!10^{-6}$. We use $\beta_1=0.9$, $\beta_2=0.95$, $\epsilon=10^{-8}$, weight decay $10^{-10}$, gradient clipping at 1, and no exponential moving average. The openpi low-memory configuration includes LoRA~\cite{lora} and other trainable modules; it updates approximately 468 million of 3.288 billion parameters. The optimization budget is matched, while recovery collection and grading incur additional computation.

\paragraph{Evaluation.}
All three policies run autonomously on the same 100 validation scenes, with ten scenes per task. Each trial restores the scene XML and initial physical state, resets policy randomness to seed 7, and allows at most 300 environment actions. The student predicts 50 actions, executes five, and then replans. Success is determined by the native environment task check. The teacher is absent during evaluation.

\subsection{$\pi_0$ results}
\label{sec:empirical_results}

\begin{figure}[t]
\centering
\includegraphics[width=0.95\linewidth]{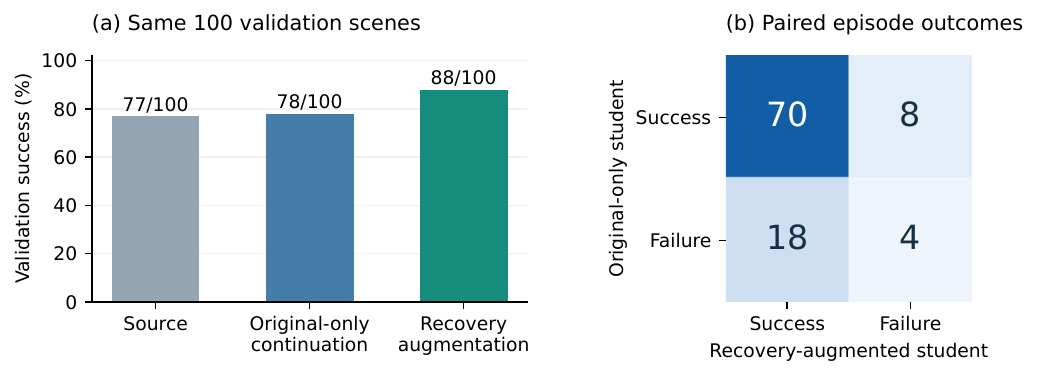}
\caption{\textbf{Observed $\pi_0$ validation performance.} (a) Successful episodes among the same 100 scenes. Both continuation conditions start from the same source and receive 2,000 updates. (b) Paired outcomes show 18 corrected failures and eight lost successes relative to original-only continuation.}
\label{fig:preliminary}
\end{figure}

\begin{table}[t]
\centering
\small
\setlength{\tabcolsep}{4pt}
\begin{tabular}{@{}p{0.52\linewidth}rrr@{}}
\toprule
Task & $M_0$ & Original & Recovery \\
\midrule
Open middle drawer & 9 & 8 & 8 \\
Open top drawer; put bowl inside & 4 & 8 & 9 \\
Push plate to stove front & 9 & 10 & 9 \\
Put bowl on plate & 5 & 8 & 10 \\
Put bowl on stove & 10 & 10 & 10 \\
Put bowl on cabinet & 10 & 8 & 10 \\
Put cream cheese in bowl & 5 & 3 & 8 \\
Put wine bottle on rack & 8 & 6 & 7 \\
Put wine bottle on cabinet & 7 & 7 & 7 \\
Turn on stove & 10 & 10 & 10 \\
\midrule
\textbf{Total successes /100} & \textbf{77} & \textbf{78} & \textbf{88} \\
\bottomrule
\end{tabular}
\caption{Preliminary $\pi_0$ results on a custom LIBERO-Goal validation split. Each task has ten trials. Original and Recovery denote 2,000 updates on original data alone and on the recovery-augmented mixture, respectively, from the same $M_0$.}
\label{tab:preliminary}
\end{table}

Recovery continuation reaches 88\% success, compared with 78\% for ordinary continuation and 77\% for $M_0$ (Table~\ref{tab:preliminary}). The gain over ordinary continuation is ten percentage points. It is not uniform across tasks: five improve, four are unchanged, and one declines. On the paired validation scenes, recovery training converts 18 ordinary-continuation failures into successes, while eight successes become failures; 70 scenes succeed under both policies and four fail under both.

The paired ordinary-versus-recovery comparison has a two-sided exact McNemar $p$-value of 0.0755, so it does not establish statistical significance at 0.05. This study uses one training seed, and its validation scenes were also used to select $M_0$.

\subsection{BC-RNN on robomimic Can}
\label{sec:bcrnn}

\paragraph{Task and policy.}
We also examine an earlier implementation of local recovery collection on the Can task from robomimic's proficient-human dataset~\cite{robomimic}. A seed-7 split assigns 50 demonstrations to training, 20 to offline validation, and the remaining 130 initial scenes to closed-loop evaluation. Their demonstration identifiers are disjoint; evaluation actions are never used for training. The original training pool contains 5,650 action labels. Each evaluation episode allows up to 400 actions at 20\,Hz and ends on the native task-success check.

The student is a custom 11.77-million-parameter BC-RNN, with an ImageNet-initialized ResNet-18 shared across two $128\!\times\!128$ RGB views. Per-view spatial-softmax features are combined with seven joint positions and two gripper positions. A one-layer LSTM with 256 hidden units uses up to ten consecutive observations, and a five-component diagonal Gaussian mixture predicts the seven-dimensional action. Training minimizes the final-frame action negative log-likelihood; inference uses the mean of the highest-weight component. The source policy is the final checkpoint after 10,000 updates, rather than the checkpoint with the best offline validation loss.

\paragraph{Local recovery data and matched continuation.}
Auditing the source policy on the 50 training initial states identifies ten failed rollouts. An auditor selects intervention states before, during, and after a failure event, and a closed-loop tool teacher generates local corrections. The retained set contains 26 segments from nine failed parent rollouts, adding 2,547 labels. It was assembled through successive collection and cropping revisions. This early teacher could access simulator geometry and physical feedback; the student receives only RGB and proprioception. The segments predate student-driven endpoint extension and the action-quality windows described in Sec.~\ref{sec:method}. They therefore test the local-supervision principle, not those later components.

Both continuation conditions load the same source checkpoint and train for 2,000 updates with batch size 48 and seed 7. They restart AdamW with a learning rate of $10^{-5}$, a 100-update warmup, weight decay $10^{-4}$, $\beta=(0.9,0.999)$, and gradient clipping at 1. Ordinary continuation samples original target frames; recovery continuation samples uniformly from the combined pool of 8,197 labels, without a fixed source quota or additional recovery weight. Up to nine preceding observations provide context, while failed student actions and subsequent student-check actions are excluded from the targets.

\begin{table}[t]
\centering
\small
\setlength{\tabcolsep}{7pt}
\begin{tabular}{@{}lrrr@{}}
\toprule
Policy & Additional updates & Successes /130 & Success (\%) \\
\midrule
Source BC-RNN & 0 & 91 & 70.00 \\
Original-data continuation & 2,000 & 102 & 78.46 \\
Recovery-augmented continuation & 2,000 & \textbf{112} & \textbf{86.15} \\
\bottomrule
\end{tabular}
\caption{BC-RNN on robomimic Can with 26 local recovery segments. Both continuation conditions start from the same source checkpoint. Results use one training seed and the same 130 evaluation initial states.}
\label{tab:bcrnn}
\end{table}

\paragraph{Results.}
With 26 recovery segments, success rises from 102/130 to 112/130 relative to ordinary continuation, a gain of 7.69 percentage points (Table~\ref{tab:bcrnn}). On the paired initial states, 16 ordinary-continuation failures become successes and six successes become failures. The additional updates alone also help: ordinary continuation improves over the source policy's 91/130. Thus the matched continuation comparison, rather than the full improvement over the source, measures the observed benefit of the recovery-augmented data pool.

\section{Protocol for Systematic Evaluation}
\label{sec:future}
The preliminary studies test whether a small amount of collected recovery supervision can improve a student under matched optimization. The following experiments are specified to separate its possible causes and assess its range. They are a prospective evaluation protocol, not completed results.

\paragraph{Independent evaluation and fair budgets.}
The first priority is to evaluate the frozen model choices on the reserved test scenes and repeat collection and training with independent seeds. All methods should share the source checkpoint, original data, student updates, and evaluation starts within each repeat. Report both nominal task completion and recovery from an independently constructed bank of failed student states. Preserve the parent rollout as the unit of data separation, and report both corrected failures and newly lost successes. Following the emphasis on uncertainty in few-run policy evaluation~\cite{agarwal2021}, report intervals that account for variation across training runs and tasks while preserving paired comparisons. Additional clean demonstrations, successful student rollouts, reactive interventions, and full-task teacher suffixes provide complementary data controls. Equal student updates do not imply equal acquisition cost: audits, replay, failed teacher attempts, student checks, scoring calls, and reset effort must be charged separately.

\paragraph{Does recursion add value?}
Compare fresh audits of each updated student with repeated training on the initial recovery pool and with a single collection round that receives the same total acquisition budget. Match total student updates and account for all collection and verification actions. After every round, evaluate autonomous task success, recovery on a fixed held-out failure bank, and retention of previously successful behavior. Track which failure types disappear, persist, or emerge. These controls test whether adapting supervision to the changing policy helps beyond additional training or a larger initial dataset.

\paragraph{Which supervision decisions matter?}
For entry selection, compare uniform points, coarse-only inspection, Focus, and Coverage on shared parent failures. Measure supported entries and unnecessary interventions as well as downstream performance. For the endpoint, compare local completion alone, progress without confirmation, fresh retries from the initial entry, and extension from the saved teacher endpoint. A length-matched control is needed to distinguish the benefit of student diagnosis from simply demonstrating more actions. Related intervention-selection principles, including expert-relative recoverability~\cite{kintsugi}, should use the same teacher interface and include their probing cost.

\paragraph{Quality and model dependence.}
Fix the accepted teacher archive when comparing whole-path imitation, quality weighting, quality-bounded intervals, and joint context-and-loss masking. Match target exposure where possible; overlapping windows must not be counted as independent demonstrations. Blind human judgments and an independent evaluator can test the utility rubric without treating the generating teacher as its own ground truth. Separately vary the teacher while holding localization, handoff checks, scoring, and tool permissions fixed. A small same-family comparison and a cross-family comparison should report data yield, retained-action precision, downstream student gain, latency, and cost. Teacher task completion alone is not a measure of teaching value.

\paragraph{Broader policies, benchmarks, and physical execution.}
Replication should apply the same complete collection procedure across policy architectures, then test stronger generalist students and more varied manipulation tasks. LIBERO supports controlled mechanism studies; RoboCasa and RoboDojo provide complementary task diversity and capability evaluations~\cite{libero,robocasa,robodojo}. Their native protocols must remain distinct from the custom validation study here. Real-robot evaluation should begin with one local repair and one student handoff from the current state, where exact rollback is not required. Multi-stage physical extension requires a separately validated restoration procedure. Hardware selection, physical results, multi-round improvement, and a best teacher model are not established by the present experiments.

\section{Discussion and Conclusion}
\label{sec:conclusion}
We formulate recursive self-improvement of visuomotor policies through local recovery supervision. Each updated student determines the next collection round through its failures and ability to continue after a repair. Within this loop, local supervision connects three decisions: recognizing an actionable failure, reaching a state the student can use, and retaining actions that are worth imitating. The proposed framework assigns these decisions to temporal evidence, frozen-student execution, and transition-level assessment. The teacher adapts its behavior within an attempt; the student acquires recovery behavior through subsequent parameter updates.

At 2,000 additional optimization steps, recovery-augmented continuation improves over original-data continuation by ten percentage points for $\pi_0$ and 7.69 points for the earlier BC-RNN pipeline. Independent evaluation, component controls, and repeated rounds are needed to establish generalization, attribute the gains, and test whether the recursive procedure sustains them. The most informative next comparison fixes the accepted trajectories while varying their training view, then fixes the training procedure while varying collection.

The method currently relies on simulation replay to restore teacher endpoints. A single-stage physical repair could avoid exact rollback, but its sensing, timing, and reset requirements remain to be validated. Multi-stage physical extension additionally requires a reliable restoration procedure. Semantic judgments also remain fallible: supported evidence and structured checks make decisions inspectable, without guaranteeing their correctness. The central premise is that demonstrations should be shaped by the student's observed failure and ability to continue, while preserving the physical correspondence between every training observation and its target actions.

\FloatBarrier
{\small\setlength{\bibsep}{2pt}
\bibliographystyle{plainnat}
\bibliography{references}}
\end{document}